\documentclass[12pt]{article}
\usepackage{times}
\usepackage{graphicx} 
\graphicspath{ {./Fig/} }
\usepackage{t1enc}
\usepackage[a4paper,left=2.54cm, right=2.54cm, top=2.54cm, bottom=2.54cm]{geometry}
\usepackage{titlesec} 
\titleformat{\section}[block]{\Large}{\thesection.}{1em}{} 
\titleformat{\subsection}[block]{\normalsize}{\thesubsection.}{1em}{} 
\titleformat{\subsubsection}[block]{\normalsize}{\thesubsubsection.}{1em}{} 
\usepackage[font=footnotesize]{caption}
\usepackage{amsmath, amsfonts, amssymb, amsthm}
\usepackage{xurl}
\usepackage[hidelinks]{hyperref}
\usepackage{physics}
\theoremstyle{definition}

\usepackage[symbol]{footmisc}

\renewcommand{\thefootnote}{\fnsymbol{footnote}}
\usepackage{multirow}
\usepackage{mathtools}
\usepackage{makecell}
\setcellgapes{3pt}

\begin{document}
\noindent Péter Antal, Tamás Péni, Roland Tóth\footnote[1]{The authors are with the Systems and Control Laboratory, Institute for Computer Science and Control, Budapest, Hungary (email: antalpeter@sztaki.hu, peni@sztaki.hu, tothroland@sztaki.hu). R. Tóth is also affiliated with the Control Systems Group of the Eindhoven University of Technology, The Netherlands.}
\vspace{12pt}\\ 
{\Large Modelling, identification and geometric control of autonomous quadcopters for agile maneuvering}
\vspace{18pt}\\
\textit{This paper presents a multi-step procedure to construct the dynamic motion model of an autonomous quadcopter, identify the model parameters, and design a model-based nonlinear trajectory tracking controller. The aim of the proposed method is to speed up the commissioning of a new quadcopter design, i.e., to enable the drone to perform agile maneuvers with high precision in the shortest time possible. After a brief introduction to the theoretical background of the modelling and control design, the steps of the proposed method are presented using the example of a 
self-developed quadcopter platform.  The performance of the method is tested and evaluated by real flight experiments.}
\vspace{12pt}

\renewcommand{\thefootnote}{\arabic{footnote}}

\section{Introduction}

The continuous development of quadcopters in recent years has led to their use in a wide range of applications, such as search and rescue missions, camera surveillance, and agricultural monitoring. As their applications continue to expand, quadcopters are expected to have increasingly complex capabilities, such as agile maneuvering in narrow spaces between obstacles, cooperating with other drones, working with humans and performing acrobatic maneuvers. The modular design of quadcopters also facilitates their wide range of applications, as the choice of components (motors, propellers, battery, sensors) can be adapted to various tasks. Agile maneuvering requires powerful motors and low weight, which typically reduces flight time. For agricultural monitoring, on the other hand, it is important that the drone can fly as long as possible on a single battery, but there is typically less need to perform fast, precise maneuvers.

In this work, the modelling, parameter identification and control of a new quadcopter with partially unknown parameters is presented in order to enable the vehicle to perform agile maneuvers with high tracking performance. An important objective is to complete the steps as quickly possible, with minimal expert knowledge, so that a new quadcopter design can be put into service in the shortest possible time. The work is focused on indoor quadcopter navigation, where an external positioning system is provided and environmental effects are limited, so that new control algorithms can be tested under ideal conditions.

To perform agile maneuvers, it is necessary to fully exploit the physical capabilities of the drone. This can only be achieved by control algorithms based on a nonlinear dynamic model of the drone that is valid over its full operating domain. The most common control methods, (PID, LQR), use a linearized model of the quadcopter around an operating point and cannot guarantee the required stability and control performance over the full state space \cite{argentim2013}, therefore more complex control strategies are needed. For the design of such algorithms, nonlinear control methods, e.g. geometric control \cite{Lee2010}, are well suited.


\section{Indoor navigation system and custom-designed quadrotor}\label{sec:setup}

\begin{figure}[!htb]
    \centering
    \includegraphics[width=13cm]{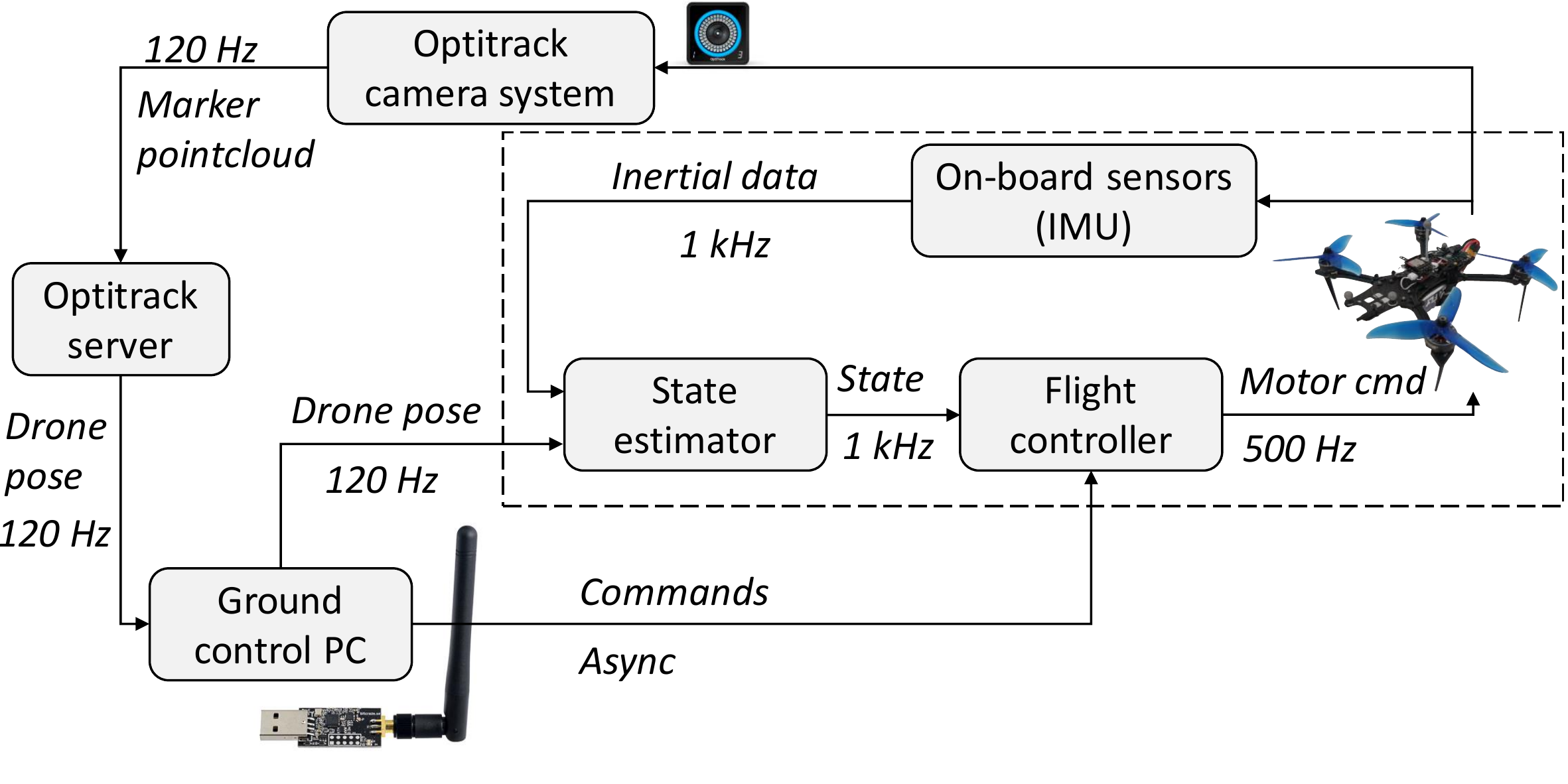}
    \caption{System architecture of the AIMotionLab test arena for indoor navigation of autonomous quadcopters.}\label{fig:system}
\end{figure}

The AIMotionLab test arena of the Systems and Control Lab of the Institute for Computer Science and Control was created to efficiently develop, implement and test path planning and motion control algorithms for various autonomous mobile robots (ground and aerial vehicles). The architecture of the indoor navigation system is shown in Figure \ref{fig:system}.

Several positioning systems can be used for indoor agile maneuvering. In many solutions, the sensor is located on the quadcopter, such as optical flow based localisation, or using on-board cameras with simultaneous localisation and mapping (SLAM). However, the reliability and accuracy of on-board sensor-based localisation is far below that of external positioning systems, such as ultra-wideband (UWB), WiFi-based, coded light-based and infrared camera-based solutions. At AIMotionLab test arena, we have chosen the infrared camera-based solution for its accuracy and reliability.

For the localization of the autonomous quadcopter, we use an Optitrack motion capture system with 14 number of Prime 13\footnote{\url{https://optitrack.com/cameras/prime-13/}} infrared cameras, which provides submillimeter accuracy in determining the position of reflective markers mounted on the quadcopter in the $3\times 3\times 2$ meter flight space. The camera images are processed by a server computer, which calculates the position and orientation of the drone, and sends it to the ground control PC. This computer communicates with the drone in real time via radio link and transmits measurements at 120 Hz together with other commands (e.g. take-off, landing, or reference trajectory). Optitrack measurements are fused with the on-board sensor measurements by a state estimator algorithm and forwarded to the motion control algorithm, which outputs PWM (pulse width modulation) signals as inputs to the motors.

The ground control computer runs the Skybrush Server\footnote{\url{https://skybrush.io/modules/server/}} drone management software with our extension\footnote{\url{https://github.com/AIMotionLab-SZTAKI/skybrush-ext-aimotionlab}}. This framework is responsible for broadcasting Optitrack measurements and high level management. The latter includes uploading reference trajectories, issuing take-off and landing commands, safety functions (through the control of certain internal states), and uploading the program running on the embedded computer. The communication between the ground control computer and the quadcopter is done via the Crazyradio\footnote{\url{https://www.bitcraze.io/products/crazyradio-pa/}} unit using CRTP\footnote{\url{https://www.bitcraze.io/documentation/repository/crazyflie-firmware/master/functional-areas/crtp/}} (Crazy RealTime Protocol).

\begin{figure}[!htb]
    \centering
    \includegraphics[width=7cm]{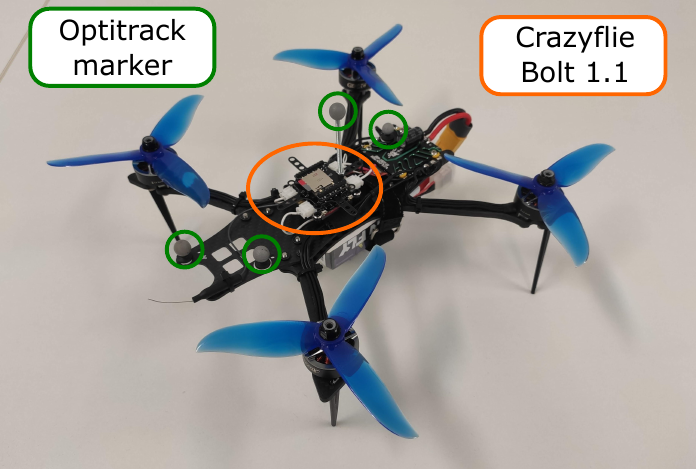}
    \caption{Bumblebee quadcopter with reflective markers.}
    \label{fig:bb}
\end{figure}

In this work, the modelling, parameter identification and control procedures are demonstrated on the example of a self-developed quadcopter named Bumblebee, shown in Figure \ref{fig:bb}. The quadcopter is based on a custom designed 250 mm wide carbon fiber frame, equipped with a 3 cell LiPo battery and 4 GEPRC GR2306.5 2450 KV BLDC motors with DALProp Dydlone T5045C Pro propellers. The components of the quadcopter have been selected to ensure that the vehicle is capable of lifting at least 250~g payload and can fly continuously for at least 10 minutes on one battery. The flight control system runs on a Crazyflie Bolt 1.1\footnote{\url{https://store.bitcraze.io/collections/kits/products/crazyflie-bolt-1-1}} embedded computer equipped with an inertial measurement unit that includes 3 DoF accelerometer and gyroscope. For the design of the quadcopter, the Crazyflie Bolt unit has been chosen because we have previously conducted research and development projects with Crazyflie 2.1 quadcopters, which are equipped with the same on-board unit (corresponding publications are e.g. \cite{Antal2022,Antal2022_2}). In addition to the inertial sensor, the Bolt unit is equipped with two microcontrollers: an STM32F405 runs the state estimation and control algorithms, and an nRF51822 handles radio communication with the ground control computer and the power distribution of the components.

\section{Quadcopter modelling}\label{sec:modelling}

\subsection{Coordinate systems}
\begin{figure}[!htb]
    \centering
    \includegraphics[width=8cm]{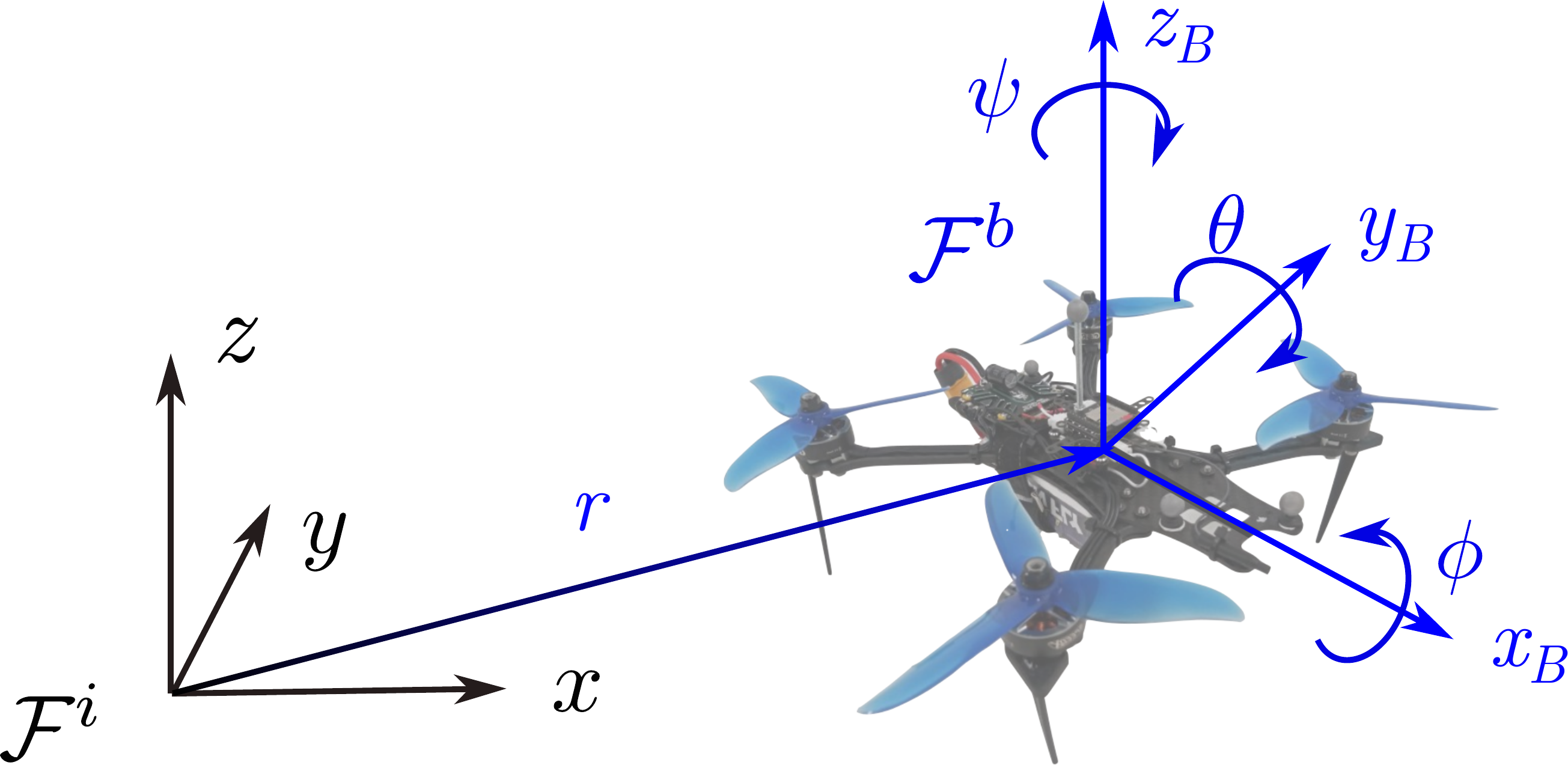}
    \caption{Inertial and body frames describing the geometric relations of the quadcopter and the environment.}
    \label{fig:model}
\end{figure}

Two coordinate systems are used to describe the motion of the quadcopter. The first is the inertial frame, denoted by $\mathcal{F}^\mathrm{i}$, which is static with respect to the environment, and the second one is the body frame, fixed to the center of mass of the quadcopter, denoted by $\mathcal{F}^\mathrm{b}$. The transformation between the two coordinate frames is described by the combination of a translation and a rotation, denoted by $r\in\mathbb{R}^3$ and $R\in\mathrm{SO}(3)$, respectively, where $\mathrm{SO}(3)$ is the group of special orthogonal matrices of size $3\times 3$. The elements of this group describe spatial rotation, and are therefore called rotation matrices, for which the following properties hold:
\begin{equation}\label{eq:rotmat}
    RR^\top = I,\quad \det(R) = 1,
\end{equation}
where $I$ is the 3-dimensional identity matrix and $\det(\cdot)$ denotes the determinant. The transformations obtained by the combination of 3D translation and rotation formulate the special Euclidean group $\mathrm{SE}(3)$, which can be interpreted as the configuration manifold of the quadcopter. An illustration of the coordinate systems and the translation vector between them are shown in Figure \ref{fig:model}.

The orientation of the quadcopter can be described in the body frame $\mathcal{F}^\mathrm{b}$ by the roll ($\phi$), pitch ($\theta$), and yaw ($\psi$), which are the \emph{Euler angles}, and their vector is denoted by $\lambda = [\phi\ \theta\ \psi]^\top \in \mathbb{R}^3$. However, it is more beneficial to use the coordinate-free rotation matrix representation for the control design discussed in Section~\ref{sec:control}. The rotation matrix from $\mathcal{F}^\mathrm{i}$ to $\mathcal{F}^\mathrm{b}$ can be calculated as a composition of successive rotations by the three Euler angles:
\begin{equation}\label{eq:rot}
\begin{split}
R_\mathrm i^\mathrm b& = \mathrm{Rot}(x,\phi)\mathrm{Rot}(z,\theta)\mathrm{Rot}(z,\psi) = \\
&=\left[\begin{array}{ccc}1 & 0 & 0 \\ 0 & C_ \phi &  S_ \phi \\ 0 & - S_ \phi & C_ \phi\end{array}\right]\left[\begin{array}{ccc} C_ \theta & 0 &  -S_ \theta \\ 0 & 1 & 0 \\  S_ \theta & 0 &  C_ \theta\end{array}\right]\left[\begin{array}{ccc} C_ \psi & S_ \psi & 0 \\ - S_ \psi &  C_ \psi & 0 \\ 0 & 0 & 1\end{array}\right]=\\
&=\left[\begin{array}{ccc} C_\psi C_\theta &  C_\theta S_\psi & -S_\theta \\
C_\psi S_\phi S_\theta - C_\phi S_\psi & C_\phi C_\psi + S_\phi S_\theta S_\psi & S_\phi C_\theta \\
S_\phi S_\psi + C_\phi C_\psi S_\theta & C_\phi S_\theta S_\psi - C_\psi S_\phi & C_\phi C_\theta
\end{array}\right],
\end{split}
\end{equation}
where $C_\phi = \cos(\phi)$, $S_\phi = \sin(\phi)$ and similarly for the other angles. Following the properties of the rotation matrix described in \eqref{eq:rotmat}, the rotation from $\mathcal{F}^\mathrm{b}$ to $\mathcal{F}^\mathrm{i}$ can be described as follows:
\begin{equation}
R_\mathrm b^\mathrm i = \left(R_\mathrm i^\mathrm b\right)^{-1} = \left(R_\mathrm i^\mathrm b\right)^\top.
\end{equation}
In the following we omit the indices of the rotation matrix and use the notation $R = R_\mathrm b^\mathrm i$ for clarity.

\subsection{Equations of motion}\label{sec:dyn_model}

\begin{figure}[!htb]
    \centering
    \includegraphics[width=9cm]{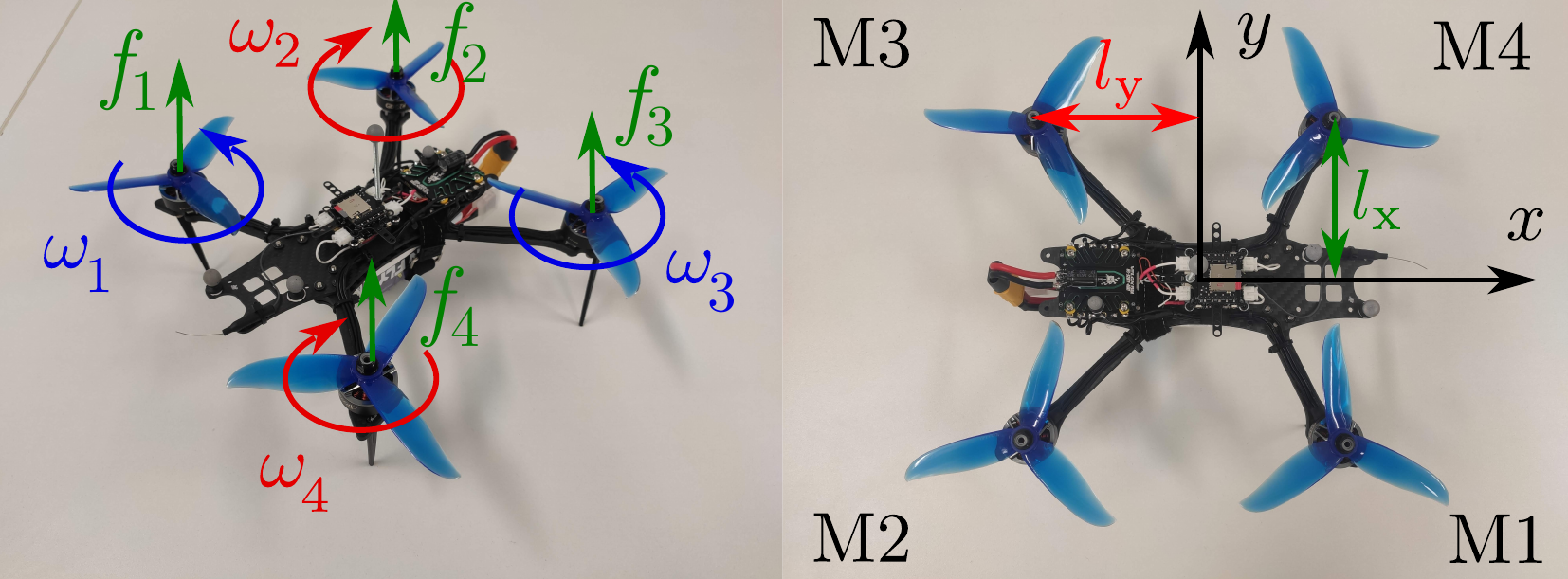}
    \caption{Bumblebee quadcopter with the directions of the propeller angular velocities and thrusts.}
    \label{fig:inputs}
\end{figure}

The quadcopter is modelled as a rigid body in 3D space, with constant mass and geometry, on which gravity acts as an external force. To achieve the desired motion, the thrust generated by the propellers is used, which is resulting from the voltage applied to the four electric motors. The relationship between the angular velocity of the propellers ($\omega_i$) and the thrust force ($f_i$) can be described as follows:
\begin{equation}\label{eq:c}
    f_i = c\omega_i^2,\quad i\in\{1,2,3,4\},
\end{equation}
where $c$ is the thrust coefficient. In addition to the thrust, each propeller generates a torque about the vertical axis of the body frame $\mathcal{F}^\mathrm{b}$, which is also described by a quadratic expression:
\begin{equation}\label{eq:b}
    \tau_{\mathrm{z}, i} = b\omega_i^2,\quad i\in\{1,2,3,4\},
\end{equation}
where $b$ is the drag coefficient. The thrust and torque generated by the four propellers can be converted into a vertical force and torques about the three axes of the body frame $\mathcal{F}^\mathrm{b}$, using the following equations:
\begin{subequations}
\begin{align}
    &F = f_1 + f_2 + f_3 + f_4,\\
    &\tau_\mathrm{x} = l_\mathrm{x}(f_3+f_4-f_1-f_2),\label{eq:taux}\\
    &\tau_\mathrm{y} = l_\mathrm{y}(f_2+f_3-f_1-f_4),\label{eq:tauy}\\
    &\tau_\mathrm{z} = b\left(\omega_1^2+\omega_3^2-\omega_2^2-\omega_4^2\right)\label{eq:tauz},
\end{align}
\end{subequations}
where $l_\mathrm{x},l_\mathrm{y}$ denote the distance of each propeller from the $x$ and $y$ axes of the body frame $\mathcal{F}^\mathrm{b}$, as it is illustrated in Figure \ref{fig:inputs}. 

The dynamic model of the quadcopter is described by the Newton--Euler equations, as follows \cite{mahony2012}:
\begin{subequations}
\begin{align}
    m\ddot{r} &= -m g e_3 + F R e_3,\label{eq:dyn1}\\
    J\dot{\omega} &= \tau - \omega \times J \omega,\label{eq:att_dyn}\\
    \dot{R} &= R \hat{\omega}\label{eq:dyn3},
\end{align}
\end{subequations}
where $m$ is the mass of the vehicle, $g$ is the gravitational acceleration, $r\in\mathbb{R}^3$ and $R\in\mathrm{SO}(3)$ are the position vector and rotation matrix introduced earlier, $e_3 = [0\ 0\ 1]^\top$, $J$ denotes the inertia matrix of the quadcopter, and $\omega$ is the angular velocity in $\mathcal{F}^\mathrm{b}$. The hat map is defined by the condition $\{\hat{(\cdot)}:\mathbb{R}^3\rightarrow \mathrm{SO}(3)\ |\ \hat{x}y=x\times y\}$, where $x,y\in\mathbb{R}^3$. The inputs of the dynamic model are the collective thrust $F\in\mathbb{R}$ and the torques about the three axes of the body frame $\tau=[\tau_\mathrm{x}\ \tau_\mathrm{y}\ \tau_\mathrm{z}]^\top\in\mathbb{R}^3$.

\subsection{Actuator model}

As mentioned in Section~\ref{sec:setup}, the output of the on-board controller for each motor is a PWM signal that determines the rate of the supply voltage applied to each motor. However, equations \eqref{eq:c}-\eqref{eq:b} express the force and torque control inputs from the angular velocity of the propellers, therefore an actuator model is required to describe the relationship between the PWM signal and the angular velocity.

Brushless DC motors are modelled in the PhD thesis \cite{Bisheban2019}, where the dynamic equations are as follows:
\begin{subequations}
\begin{align}
    &J_\mathrm{m} \dot \omega_\mathrm{m} = -k_\mathrm{m} \omega_\mathrm{m} + \tau_\mathrm{e} - \tau_\mathrm{m},\label{eq:mot_dyn_1}\\
    &\tau_\mathrm{e} = k_\mathrm{q,0}i_\mathrm{a} - k_\mathrm{q,1}i_\mathrm{a}^2,\label{eq:mot_dyn_2}\\
    &U = k_\mathrm{e} \omega_\mathrm{m} + R_\mathrm{a} i_\mathrm{a},\label{eq:mot_dyn_3}
\end{align}
\end{subequations}
where $J_\mathrm{m}$ is the moment of inertia of the rotor, $k_\mathrm{m}, k_\mathrm{e}, k_\mathrm{q,0}, k_\mathrm{q,1}$ are constants, $R_\mathrm{a}$ is the armature resistance, $i_\mathrm{a}$ is the armature current, $\tau_\mathrm{e}$ is the induced torque, $\tau_\mathrm{m}$ is the load torque, and $U$ is the input voltage, calculated from the PWM value and the maximum supply voltage: $U = U_\mathrm{max}\mathrm{PWM}$. Here, the angular velocity of the rotor is denoted by $\omega_\mathrm{m}$, since, unlike in the previous chapter, the equations describe the behaviour of only one motor. However, in the following chapters, similarly to the previous ones, the numbering of each motor is indicated in the subscript ($\omega_i, i\in\{1,2,3,4\}$).

The dynamics described by \eqref{eq:mot_dyn_1} are very fast compared to the change of the input in case of small BLDC motors, therefore we neglect it. The equations \eqref{eq:mot_dyn_2}-\eqref{eq:mot_dyn_3} describe a static relation between the input voltage (hence PWM) and the motor angular velocity, which can be expressed in general as follows:
\begin{equation}\label{eq:pwm-ang-generic}
    \omega_\mathrm{m} = f_\mathrm{m}(\mathrm{PWM}, \vartheta).
\end{equation}
The realization of the static characteristics $f_\mathrm{m}$ and the definition of its parameters, denoted by $\vartheta$, are described in detail later in Section~\ref{sec:pwm-ang_vel}. All parameters of the quadcopter model -- the identification of which is necessary to implement the control algorithms -- are given in Table~\ref{tab:model_param}.

\begin{table}[!h]
    \centering \footnotesize
    \begin{tabular}{c|c|c}
      Parameter   & Notation & Unit \\ \hline \hline
      Mass   & $m$ & kg \\ \hline
      Inertia matrix & $J$ & kgm$^2$ \\ \hline
      Propeller distance & $l_\mathrm{x}, l_\mathrm{y}$ & m \\ \hline
      Drag coefficient & $b$ & Nms$^2$/rad \\ \hline
      Thrust coefficient & $c$ & Ns$^2$/rad \\ \hline
      Actuator parameters & $\vartheta$ & -
    \end{tabular}
    \caption{The parameters of the quadcopter model, which are unknown for a new design.}
    \label{tab:model_param}
\end{table}

\section{Geometric control}\label{sec:control}

\subsection{Overview}
The dynamics of quadcopters, as introduced in the previous chapter, are inherently unstable, therefore a low-level controller is needed to both stabilize the vehicle by adjusting the motor voltages and to ensure that it follows the reference trajectory. The low-level control of quadcopters is an extensively researched topic with several results. The simplest and most widespread solution is the use of linear time invariant (LTI) controllers, such as PID and LQR \cite{argentim2013}. In the design of these control algorithms, the introduced nonlinear equations of motion are linearized around an operating point (typically the hovering state), and then a controller can be systematically constructed and its parameters tuned for this LTI system. A slightly more complex method is feedback linearization, where the differential flatness property of the dynamic model is exploited to transform the nonlinear system into a linear one by state feedback, and thus also apply the LTI control design process \cite{lee2009}.

The low-level control algorithms mentioned so far are relatively easy and fast to implement, however, these do not guarantee stability over the entire operating domain of the quadcopter and are mostly unable to accurately track fast, agile reference trajectories. To overcome these limitations, a geometric controller is presented in this chapter, based on \cite{Lee2010}. The geometric controller design is performed directly on the configuration manifold of the quadcopter ($\mathrm{SE}(3)$), resulting in exponential convergence of the tracking errors over the full operating domain, thus providing the performance required for high-speed, agile maneuvering.

\subsection{Differential flatness}\label{sec:flatness}
The dynamic model presented in Section~\ref{sec:dyn_model} is underactuated, since it has 6 degrees of freedom and only 4 inputs. However, in control design, we exploit the fact that the model is \textit{differentially flat}, which means that all states and control inputs can be expressed in terms of the time derivatives of the \emph{flat outputs} \cite{nieuwstadt1996}. In case of the model described by \eqref{eq:dyn1}-\eqref{eq:dyn3}, the flat outputs are the yaw angle and the 3 coordinates of the position vector (the derivation can be found, among others, in \cite{mellinger2011}). This means, that it is sufficient to design reference trajectories $x_\mathrm d(t), y_\mathrm d(t), z_\mathrm d(t), \psi_\mathrm d(t)$, which uniquely determine the trajectories of all other state variables.

\subsection{Controller synthesis}
\begin{figure}
    \centering
    \includegraphics[width=5cm]{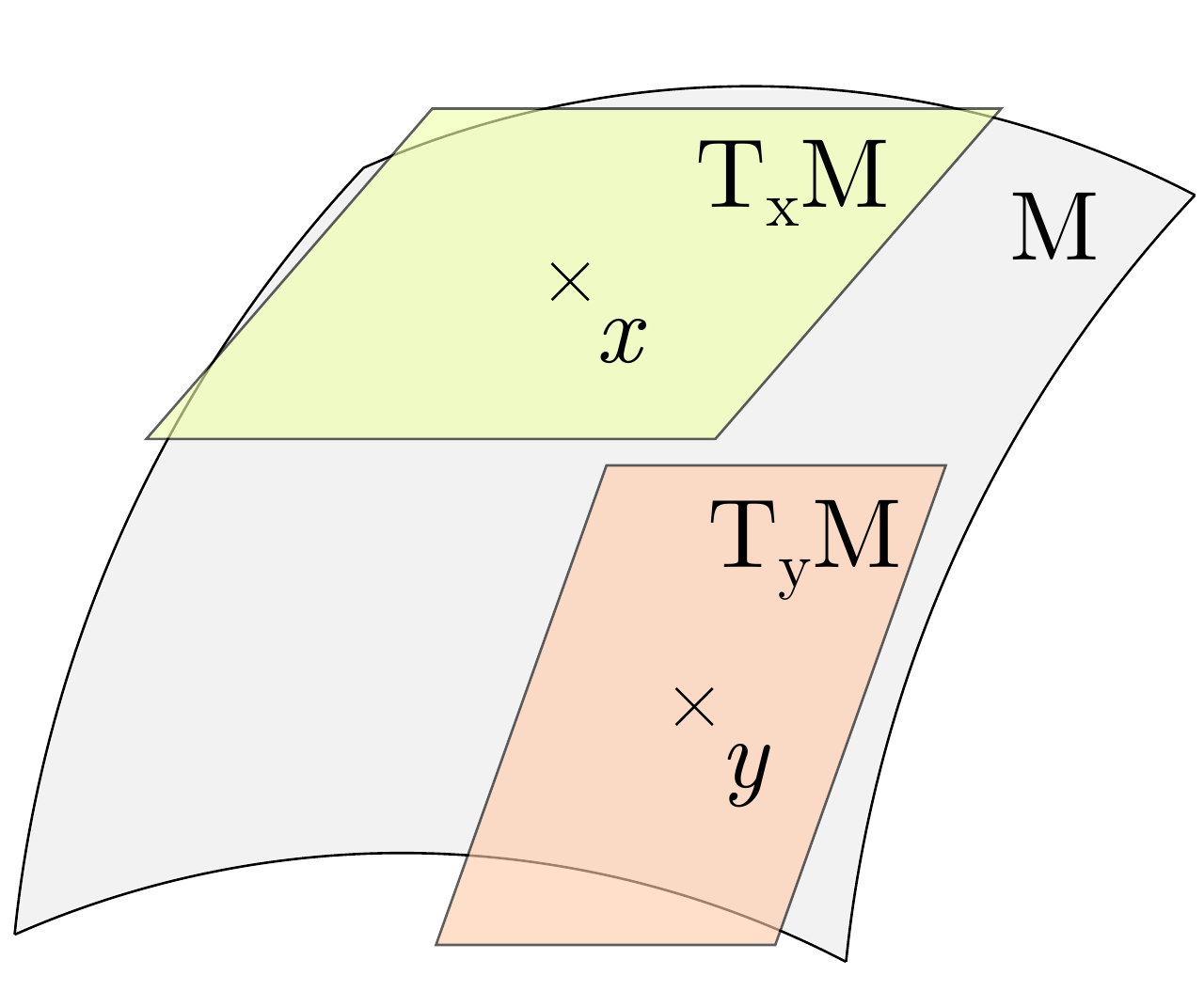}
    \caption{Visual representation of a manifold M with two tangent spaces: $\mathrm{T_x M}$ at point $x$ and $\mathrm{T_y M}$ at point $y$. It is clear that a vector in $\mathrm{T_x M}$ and another in $\mathrm{T_y M}$ can not be compared unless we transform them to the same space.}
    \label{fig:mani}
\end{figure}

The aim of the geometric controller is to track reference position and orientation, denoted by $r_\mathrm{d}(t) = [x_\mathrm{d}(t), y_\mathrm{d}(t), z_\mathrm{d}(t)]^\top $, and $R_\mathrm{d}(t)\in \mathrm{SO}(3)$, respectively, but the explicit notation of time dependence is neglected in the derivation for clarity. The position and velocity tracking errors are defined as the difference between the state and the reference signal:
\begin{align}
    &e_\mathrm{r} = r-r_\mathrm{d},\\
    &e_\mathrm{v} = v-v_\mathrm{d}.
\end{align}
For the orientation and angular velocity, we express the error terms such that they evolve on the tangent space of the configuration manifold (which in this case is $\mathrm{SO}(3)$) at each time instant. For this, we introduce the following attitude error function \cite{Lee2010}:
\begin{equation}\label{eq:psi}
    \Psi(R,R_\mathrm{d}) = \frac{1}{2}\mathrm{tr}\left( I-R_\mathrm{d}^\top R\right),
\end{equation}
where $\mathrm{tr}(\cdot)$ is the trace operator and $I$ is the identity matrix of size $3\times 3$. The error function is locally positive definite around $R_\mathrm{d}^\top R = I$ within the region where the angle of the current and reference orientations is less than $\pi$. Using the identity $-\frac{1}{2}\mathrm{tr}(\hat{x}\hat{y})=x^\top y$, the derivative of the error function is given as follows:
\begin{align}
    \mathrm{D}_\mathrm{R}\Psi(R,R_\mathrm{d})\cdot R \hat{\eta}=\frac{1}{2}\left(R_\mathrm{d}^\top R - R^\top R_\mathrm{d}\right)^\vee\cdot\eta,
\end{align}
where the variation of the rotation matrix is expressed as $\delta R=R\hat{\eta}, \eta\in\mathbb{R}^3$, and the \emph{vee operator} $(\cdot)^\vee:\mathrm{SO}(3)\rightarrow \mathbb{R}^3$ is the inverse of the hat operator $(\hat{\cdot})$ introduced previously. From this, the attitude tracking error is defined as
\begin{equation}
    e_\mathrm{R} = \frac{1}{2}\left(R_\mathrm{d}^\top R - R^\top R_\mathrm{d}\right)^\vee.
\end{equation}
The derivative of the current and reference rotations $\dot{R}\in\mathrm{T}_\mathrm{R}\mathrm{SO}(3)$ and $\dot{R}_\mathrm{d}\in\mathrm{T}_{\mathrm{R}_\mathrm{d}} \mathrm{SO}(3)$ lie in different tangent spaces (see the illustration in Figure~\ref{fig:mani}), therefore they cannot be directly compared to calculate the angular velocity error. This is addressed by transforming $\dot{R}_\mathrm{d}$ into a vector in $\mathrm{T}_\mathrm{R}\mathrm{SO}(3)$, and comparing it with $\dot{R}$, as follows:
\begin{equation}
    \dot{R} - \dot{R}_\mathrm{d}(R_\mathrm{d}^\top R)=R\hat{\omega}-R_\mathrm{d}\hat{\omega}_\mathrm{d}R_\mathrm{d}^\top R = R(\omega-R^\top R_\mathrm{d}\omega_\mathrm{d})^\wedge.
\end{equation}
The angular velocity tracking error can be now defined as
\begin{equation}
    e_\omega = \omega-R^\top R_\mathrm{d}\omega_\mathrm{d},
\end{equation}
which is the angular velocity of the rotation matrix $R_\mathrm{d}^\top R$ represented in the body-fixed frame.

\begin{figure}
    \centering
    \includegraphics[width=.6\linewidth]{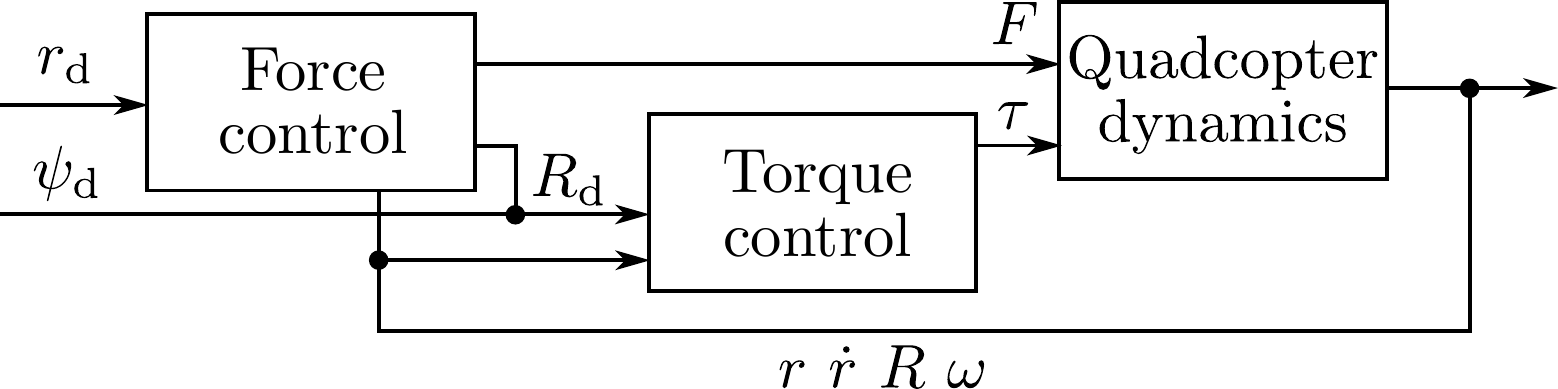}
    \caption{Block diagram of the geometric feedback control.\cite{Lee2010}}
    \label{fig:geom_scheme}
\end{figure}

Using the tracking error terms, the thrust and torque inputs introduced in Section~\ref{sec:dyn_model} are regulated by the following control law:
    \begin{align}
        F =&  (-k_\mathrm{r}e_\mathrm{r} - k_\mathrm{v}e_\mathrm{v} - mge_3 + m\ddot{r}_\mathrm{d})^\top R e_3,\label{eq:geomforce}\\
            \tau =&  -k_\mathrm{R} e_\mathrm{R} - k_\omega e_\omega + \omega \times J\omega -J\left(\hat{\omega} R^\top R_{\mathrm{d}} \omega_{\mathrm{d}}-R^\top R_{\mathrm{d}} \dot{\omega}_{\mathrm{d}}\right),\label{eq:geomtau}
      \end{align}
where $k_\mathrm{r}, k_\mathrm{v}, k_\mathrm{R}, k_\omega \in \mathbb{R}$ are the gains of the model based controller. The resulting geometric controller \cite{Lee2010} ensures that the zero equilibrium of the tracking errors of the complete dynamics is exponentially stable, provided that the reference trajectory satisfies the following requirements:
\begin{equation}
    \|-mge_3+m\ddot{r}_\mathrm{d}\| < B
\end{equation}
for a given positive constant $B$, and the initial conditions satisfy
\begin{align}
    \begin{split}
      &\Psi(R(0), R_\mathrm{d}(0))  \leq \psi_1 < 1,\\
    & \left\|e_{\omega}(0)\right\|^{2}<\frac{2}{\lambda_{\min }(J)} k_\mathrm{R}\left(1-\Psi\left(R(0), R_\mathrm{d}(0)\right)\right)      
    \end{split}
\end{align}
where $\psi_1$ is a user-defined constant, and $\lambda_{\min}(J)$ denotes the smallest eigenvalue of the inertia matrix. The stability is proven in \cite{Lee2010} by constructing positive definite Lyapunov functions for the attitude and position dynamics separately, and ensuring that the time derivative of the Lyapunov candidates are negative within the stable region of the control gains.

As mentioned in Section~\ref{sec:flatness}, the dynamic model of the quadcopter is differentially flat, therefore it is sufficient to design a reference trajectory for the position and the yaw angle. The reference rotation matrix used in the control algorithm $R_\mathrm{d}=[r_1, r_2, r_3]$ can be expressed from the flat outputs as follows:
\begin{subequations}
\begin{align}
    r_1 &= r_2\times r_3,\\
    r_2 &= \frac{r_3\times [\cos\psi_\mathrm{d}, \sin\psi_\mathrm{d}, 0]}{\norm{r_3\times [\cos\psi_\mathrm{d}, \sin\psi_\mathrm{d}, 0]}},\\
    r_3 &= \frac{-k_\mathrm{r} e_\mathrm{r} - k_\mathrm{v} e_\mathrm{v} + m g e_3+m \ddot r_\mathrm{d}}{\norm{-k_\mathrm{r} e_\mathrm{r} - k_\mathrm{v} e_\mathrm{v} + m g e_3+m \ddot r_\mathrm{d}}}.
\end{align}
\end{subequations}
The third column vector of the reference rotation matrix, $r_3$, always points in the direction of the required thrust input, $r_2$ is always perpendicular to the other two column vectors, and $r_1$ expresses the heading of the vehicle. 
\section{Parameter identification}
The geometric control algorithm presented in the previous chapter depends on the physical parameters of the actual quadcopter design, hence stability and required control performance can only be guaranteed if the numerical values of these parameters are known. The parameters that need to be identified are collected in Table \ref{tab:model_param}.

\subsection{Mass and inertia}
The mass of the vehicle can be measured using a simple scale, and the obtained numerical value is $m=0.605\ \mathrm{kg}$ for the Bumblebee quadrotor. However, determining the inertia matrix is more complicated, as direct measurement requires dynamic tests. Several examples of such measurement setups can be found in the literature (see \cite{Forster, Mustapa2014}), but they require special equipment and measurement tools. Another common method is the construction of a CAD model of the vehicle, as modern CAD software is able to calculate the inertia of the assembled rigid body based on the physical parameters of the individual components. We have chosen the latter method, since it is faster and more cost-effective to implement, and the CAD model can later be used for simulations, as well.

\begin{figure}[!htb]
    \centering
    \scalebox{-1}[1]{\includegraphics[height=4cm]{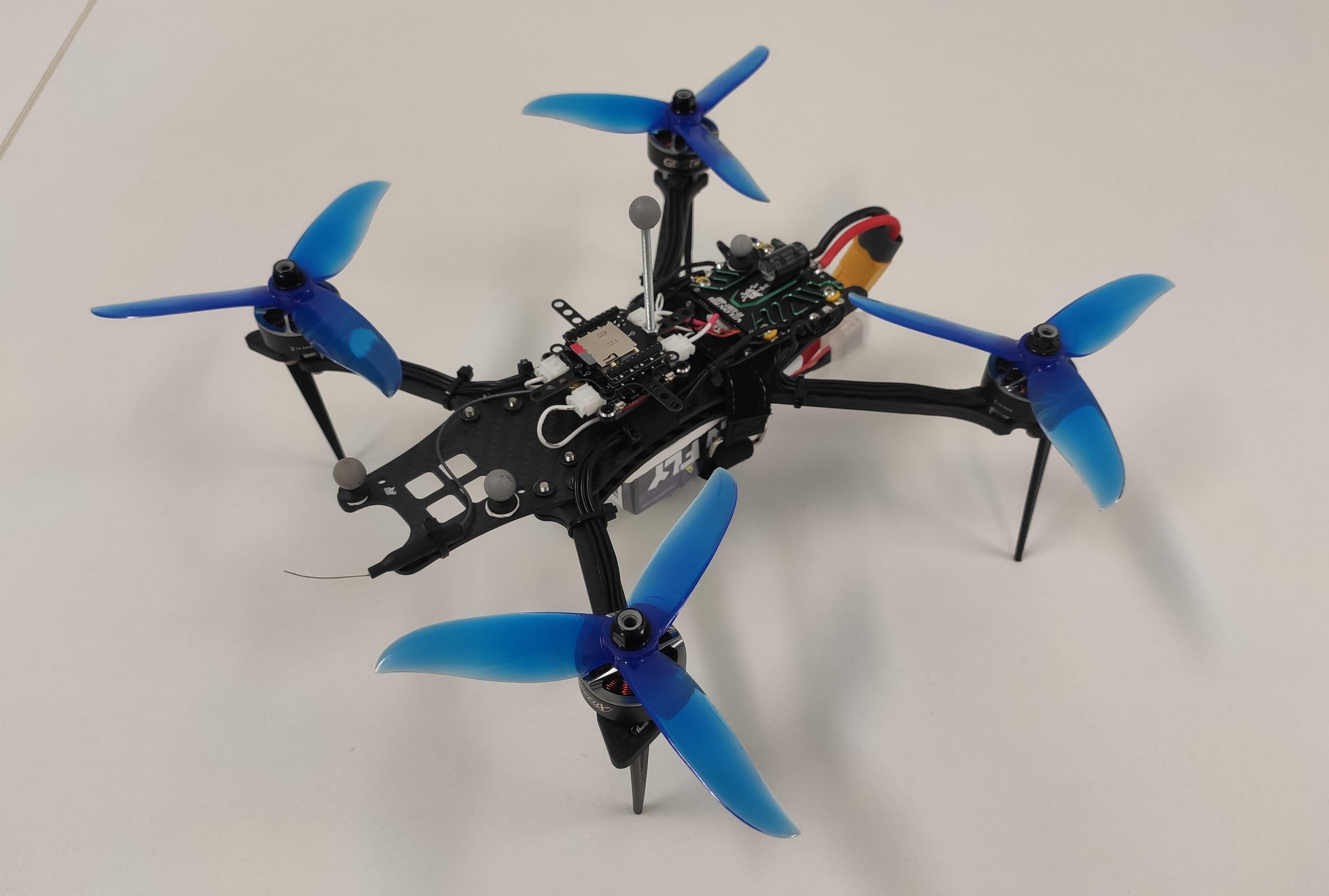}}
    \includegraphics[height=4cm]{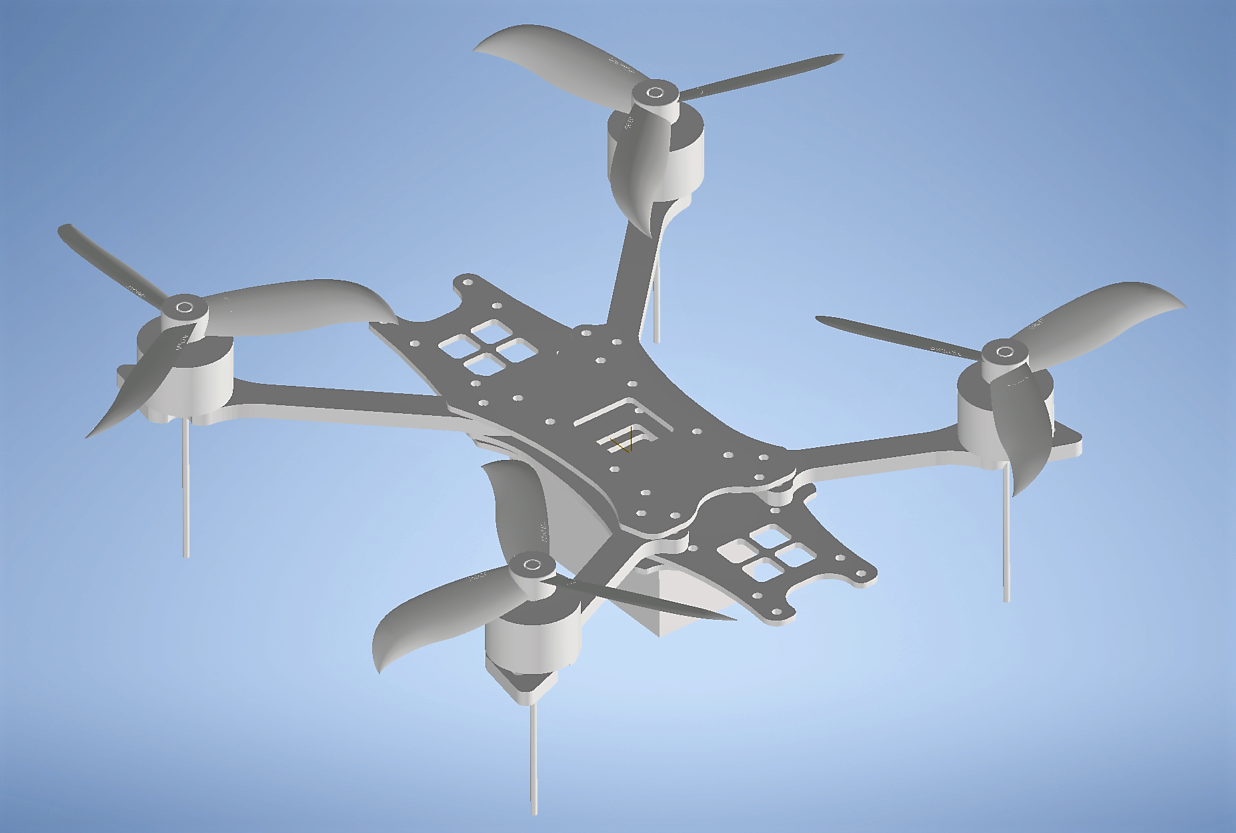}
    \caption{The Bumblebee quadcopter and the designed CAD model.}
    \label{fig:cad}
\end{figure}

The 3D model is shown in Figure \ref{fig:cad} along with the real quadcopter. The exact geometry and mass of most components are known, such as the carbon fiber frame, the propellers and the battery, thus they have been included in the model directly. The BLDC motors have been approximated with a cylinder geometry and their mass is also known. The 3D model does not include the cables and the reflective markers used for Optitrack motion tracking, but their mass is negligible, thus they have very small effect on the inertia. Once the dimensions and mass (density) of the components have been specified, the CAD software is able to calculate the inertia matrix, the obtained numerical value of which is as follows:
\begin{equation}
    J_\mathrm{CAD} = \begin{bmatrix}
        J_\mathrm{xx} & J_\mathrm{xy} & J_\mathrm{xz} \\ J_\mathrm{yx} & J_\mathrm{yy} & J_\mathrm{yz} \\ J_\mathrm{zx} & J_\mathrm{zy} & J_\mathrm{zz}
    \end{bmatrix} = \begin{bmatrix}
        1590.5 & -3.14 & -0.135 \\ -3.14 & 1481.3 & -49.18 \\ -0.135 & -49.18 & 2768.4
    \end{bmatrix}\ \mathrm{kg\ mm}^2.
\end{equation}
Since the geometry of the vehicle is nearly symmetric, the diagonal elements are 2-3 orders of magnitude larger than the others, thus the latter are often approximated by 0.

\subsection{Input parameters}

The input of the quadcopter dynamic model is $[F\ \tau^\top]^\top$, as described in Section~\ref{sec:dyn_model}. In contrast, as mentioned in Section~\ref{sec:setup}, the motors are controlled by PWM signal, which is a fill factor, i.e., the rate of the supply voltage that is applied to each motor. Therefore, to implement the control algorithm given by \eqref{eq:geomforce}-\eqref{eq:geomtau} on the real system, it is necessary to define the PWM--thrust and PWM--torque relations. These relations are partially given by equations \eqref{eq:c}-\eqref{eq:b}, but then it is still necessary to measure the thrust coefficient and the drag coefficient, moreover, to obtain the relationship between the PWM signal and the motor angular velocity. These characteristics are discussed in the following.

\subsubsection{PWM--angular velocity}\label{sec:pwm-ang_vel}
We describe the PWM--angular velocity characteristics with \eqref{eq:pwm-ang-generic} using the parameters $\vartheta=[\vartheta_1\ \vartheta_2]^\top$. Similarly to determining the inertia matrix, the aim was to use the simplest possible methods and the few sensors for the measurements, in order to estimate the parameters of a new quadcopter design quickly and easily. In the current case, an external sensor (e.g. a tachometer) could be used to measure the angular velocity of the motor, however, it is possible to estimate it without additional equipment. The data sheet of most BLDC motors for quadcopters include a parameter KV, which indicates the increase in motor rpm for a voltage increase of 1 V (volt). The data sheet for the motor we are using shows 2450 KV\footnote{\url{https://geprc.com/product/geprc-gr2306_5-1350kv-1850kv-2450kv-motors/}}.

In addition, the PWM--angular velocity characteristics are influenced by the fact that the motor does not start rotating below a certain threshold $\mathrm{PWM}_0$, therefore the linear characteristics are only valid above this threshold. To determine the numerical value of the threshold, we increased the PWM signal until the motor started rotating, which is $\mathrm{PWM}_0=0.0305$. This gives an estimate of the characteristics:
\begin{align}
    \omega_i &= \vartheta_1 + \vartheta_2 \cdot \mathrm{PWM}_i,\label{eq:ang_vel_pwm}\\
    \frac{\partial \omega_i }{\partial U_i} &= 2450\;\frac{\mathrm{rpm}}{\mathrm{V}},\\
    \frac{\partial \omega_i }{\partial \mathrm{PWM}_i} &= 2450 \cdot \frac{2\pi}{60} \cdot \frac{U_\mathrm{max}}{\mathrm{PWM}_\mathrm{max}} = 2450 \cdot \frac{2\pi}{60}\cdot \frac{16.8}{1} \; \frac{\mathrm{rad}}{\mathrm{s}} = 4310.17 \; \frac{\mathrm{rad}}{\mathrm{s}} = \vartheta_2,\label{eq:c2}\\
    \vartheta_1 &= -\mathrm{PWM}_0 \cdot \vartheta_2 = -0.0305 \cdot \vartheta_2 = -131.538\; \frac{\mathrm{rad}}{\mathrm{s}},\label{eq:c1}\\
    \omega_i &= -131.538 + 4310.17 \cdot \mathrm{PWM}_i,\label{eq:pwm-to-angvel}
\end{align}
where $i\in\{1,2,3,4\}$ is the index of the motor, $\mathrm{PWM}_i \in [0, 1]$ is the PWM signal, $\omega_i$ is the angular velocity of the motor, and $U_\mathrm{max}=16.8$ V is the maximum supply voltage.

\subsubsection{Angular velocity--thrust}
The measurement of angular velocity--thrust characteristics gives an estimate of the amount of thrust generated by the propellers rotating with a specific angular velocity. Two measurement procedures were used to determine the relationship:
\begin{enumerate}
    \item Equipped with different weights, the PWM signals were recorded in hovering, and the thrust is then calculated as $f_i = mg/4$ for all propellers.
    \item The quadcopter was put on a scale and uniform PWM signals were applied to the motors such that the vehicle did not take off, and weight measured by the scale was recorded. The thrust force is then $f_i = (m - m_s) g / 4$, where $m_s$ is the mass value displayed by the scale.
\end{enumerate}
The PWM signals were converted to angular velocity according to \eqref{eq:pwm-to-angvel}. It is important to note that method 1. requires a controller to enable the drone to hover in a constant position. For this, we used a PID controller, which is available as a built-in feature on the Crazyflie Bolt~1.1 on-board computer\footnote{\url{https://www.bitcraze.io/documentation/repository/crazyflie-firmware/master/functional-areas/sensor-to-control/controllers/}}. However, although this controller allows the drone to hover, it is not suitable for fast maneuvers.
\begin{figure}[!htb]
    \centering
    \includegraphics[width=12cm]{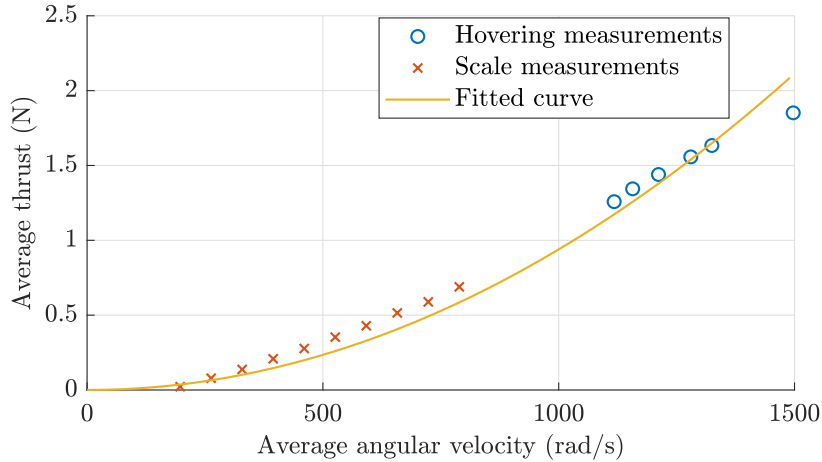}
    \caption{PWM–thrust diagram: The quadratic curve provides a good approximation and
fits to the model described by \eqref{eq:c}.}
    \label{fig:angvel-to-thrust}
\end{figure}

The measurement results are shown in Figure \ref{fig:angvel-to-thrust}. Average values are displayed on the axes, since during hovering the instantaneous PWM signals output by the controller may differ for each motor, but on average they only compensate for gravity. In yellow is the fitted curve based on \eqref{eq:c}, which is expressed as follows:
\begin{align}
    f_i = c \omega_i^2 = 9.3945\cdot 10^{-7}\ \frac{\mathrm{Ns}^2}{\mathrm{rad}},
\end{align}
from which the numerical value of the thrust coefficient is $c=9.3945\cdot 10^{-7}$ Ns$^2$/rad. Then, we can express the PWM--thrust characteristics:
\begin{align}
    f_i &= c (\vartheta_1 + \vartheta_2 \mathrm{PWM}_i)^2 = c \vartheta_1^2 + 2 c  \vartheta_1 \vartheta_2 \mathrm{PWM}_i + c \vartheta_2^2 \mathrm{PWM}_i^2 \\
    f_i &= 0.0163 - 1.065\ \mathrm{PWM}_i + 17.453\ \mathrm{PWM}_i^2 \;\ \mathrm{N}\label{eq:pwm-to-thrust},
\end{align}
where the numerical value of $\vartheta_1$ and $\vartheta_2$ are substituted from \eqref{eq:c2}-\eqref{eq:c1}.

\subsubsection{Angular velocity--torque}
The only remaining input parameter is the drag coefficient, which describes the angular velocity--torque characteristics, nominally given by \eqref{eq:b}. The third component of \eqref{eq:att_dyn} is used for the measurement, written as follows:
\begin{equation}\label{eq:att_z_dyn}
    J_\mathrm{zz} \dot\omega_\mathrm{z} = \tau_\mathrm{z} - (J_\mathrm{yx}\omega_\mathrm{x}^2 + J_\mathrm{yy}\omega_\mathrm{x}\omega_\mathrm{y} + J_\mathrm{yz}\omega_\mathrm{x}\omega_\mathrm{z} -  J_\mathrm{xx}\omega_\mathrm{x}\omega_\mathrm{y} - J_\mathrm{xy}\omega_\mathrm{y}^2 - J_\mathrm{xz}\omega_\mathrm{y}\omega_\mathrm{z}).
\end{equation}
During the measurement, the reference position of the quadcopter is kept constant and the reference orientation is set to the following trajectory:
\begin{align}\label{eq:att_ref}
    \phi_\mathrm{d}(t) = 0,\quad \theta_\mathrm{d}(t) = 0,\quad  \psi_\mathrm{d}(t) = \begin{cases}
        \frac{1}{2}\alpha t^2 & t\in \left[0, \frac{T}{2}\right) \\
        -\frac{1}{2}\alpha (t-T)^2 & t\in \left[\frac{T}{2}, T\right]
    \end{cases}.
\end{align}
It is easy to see that the absolute value of the reference acceleration of the yaw angle given in this form is constant, $|\ddot{\psi}_\mathrm{d}(t)|=\alpha$. Assuming that the reference orientation given by \eqref{eq:att_ref} is followed precisely, the parenthesized term in \eqref{eq:att_z_dyn} is zero and the left and right sides of the equation are piecewise constant, only changing sign halfway through the measurement interval. Performing the simplifications and substituting \eqref{eq:tauz} results in the following equation:
\begin{equation}\label{eq:torque_nom}
    J_\mathrm{zz} \dot\omega_\mathrm{z} =\tau_\mathrm{z} = b\left(\omega_1^2+\omega_3^2-\omega_2^2-\omega_4^2\right),
\end{equation}
where $J_\mathrm{zz}$ is previously determined, $\dot\omega_\mathrm{z}$ can be measured by numeric differentiation of the quadcopter's on-board gyroscope signal, and the propeller angular velocities can be calculated from the motor PWM signals using \eqref{eq:pwm-to-angvel}.

\begin{figure}[!htb]
    \centering
    \includegraphics[width=12cm]{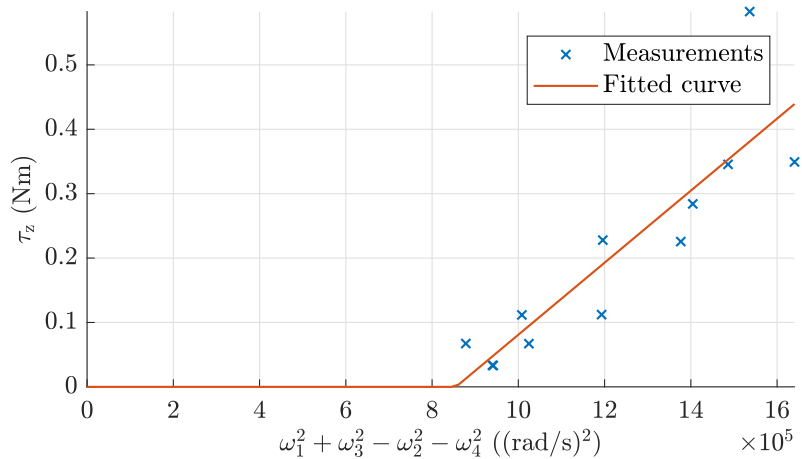}
    \caption{PWM--torque diagram: the characteristics described by \eqref{eq:torque_nom} need to be extended with an offset value.}
    \label{fig:angvel-to-torque}
\end{figure}

The measurement results are shown in Figure \ref{fig:angvel-to-torque}, where different points correspond to different $\alpha$ values at $\psi_\mathrm{d}(t)$. From the figure it is clear that in this case expression \eqref{eq:torque_nom} does not fit the measurements well enough, however, by adding a constant offset, we can describe the behaviour accurately. The resulting characteristics are as follows:
\begin{equation}
   J_\mathrm{zz} \dot\omega_\mathrm{z} = \tau_\mathrm{z} = b_1\left(\omega_1^2+\omega_3^2-\omega_2^2-\omega_4^2\right) + b_2,
\end{equation}
where the numerical values of the coefficients are $b_1 = 5.5939 \cdot 10^{-7}$ and $b_2 = - 0.4785$. This provides all the physical parameters required for implementation of the model-based control. The tools and algorithms used allow the application of the method to other quadcopters (e.g. smaller and larger), which confirms the practical efficiency of the presented methods.
\section{Control parameter tuning} 

In addition to the model parameters, the control performance also depends on the constants introduced in equations \eqref{eq:geomforce}, \eqref{eq:geomtau}. Although it is possible to check whether the closed loop is stable for given parameter values based on the literature, there is no instruction for the selection of control gains that maximize the closed loop performance with respect to given performance specification (e.g. minimizing tracking error). Therefore, the controller parameters are tuned based on flight experiments.

As shown in Figure \ref{fig:geom_scheme}, the geometric controller has a cascade structure, where the inner loop is the attitude (torque) control and the outer loop is the position (force) control. Due to the cascade structure, the outer loop can only be tuned if the inner loop is already stable. Hence, the attitude control was tuned first, followed by the position control.

\subsection{Attitude control tuning}

\begin{figure}[!htb]
    \centering
    \includegraphics[width=15cm]{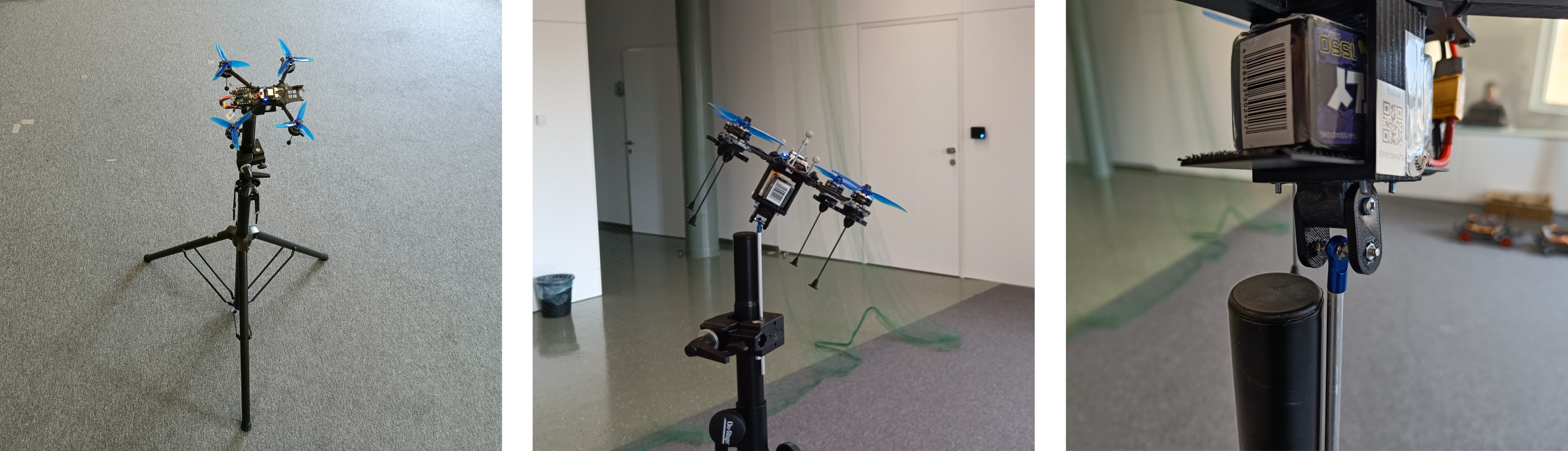}
    \caption{Measurement setup to determine the attitude control parameters.}
    \label{fig:att_tun}
\end{figure}

The attitude control law is given by \eqref{eq:geomtau}, where the first two terms implement feedback and the second two implement feedforward. The tunable parameters are the coefficients of the feedback terms, i.e. $k_\mathrm{R}, k_\omega \in \mathbb{R}$. Similar to the procedure presented in the PhD thesis \cite{Bisheban2019}, we mounted the quadcopter to a spherical joint in order to decouple the rotational and translational subsystems. The measurement setup is shown in Figure \ref{fig:att_tun}: the spherical joint is attached to the stand, to which the quadcopter is connected via a self-designed 3D printed element.

The designed layout allows to safely test the controller for different parameter settings. After evaluating several combinations of $k_\mathrm{R}, k_\omega$, we obtained that the following parameters empirically maximize the closed loop performance:
\begin{equation}
    k_\mathrm{R}=0.6,\quad k_\omega=0.15.
\end{equation}
A video of the flight experiments is available at \url{https://youtu.be/9r_j21rd800}.

\subsection{Position control tuning}

\begin{figure}[!b]
    \centering
    \includegraphics[width=12cm]{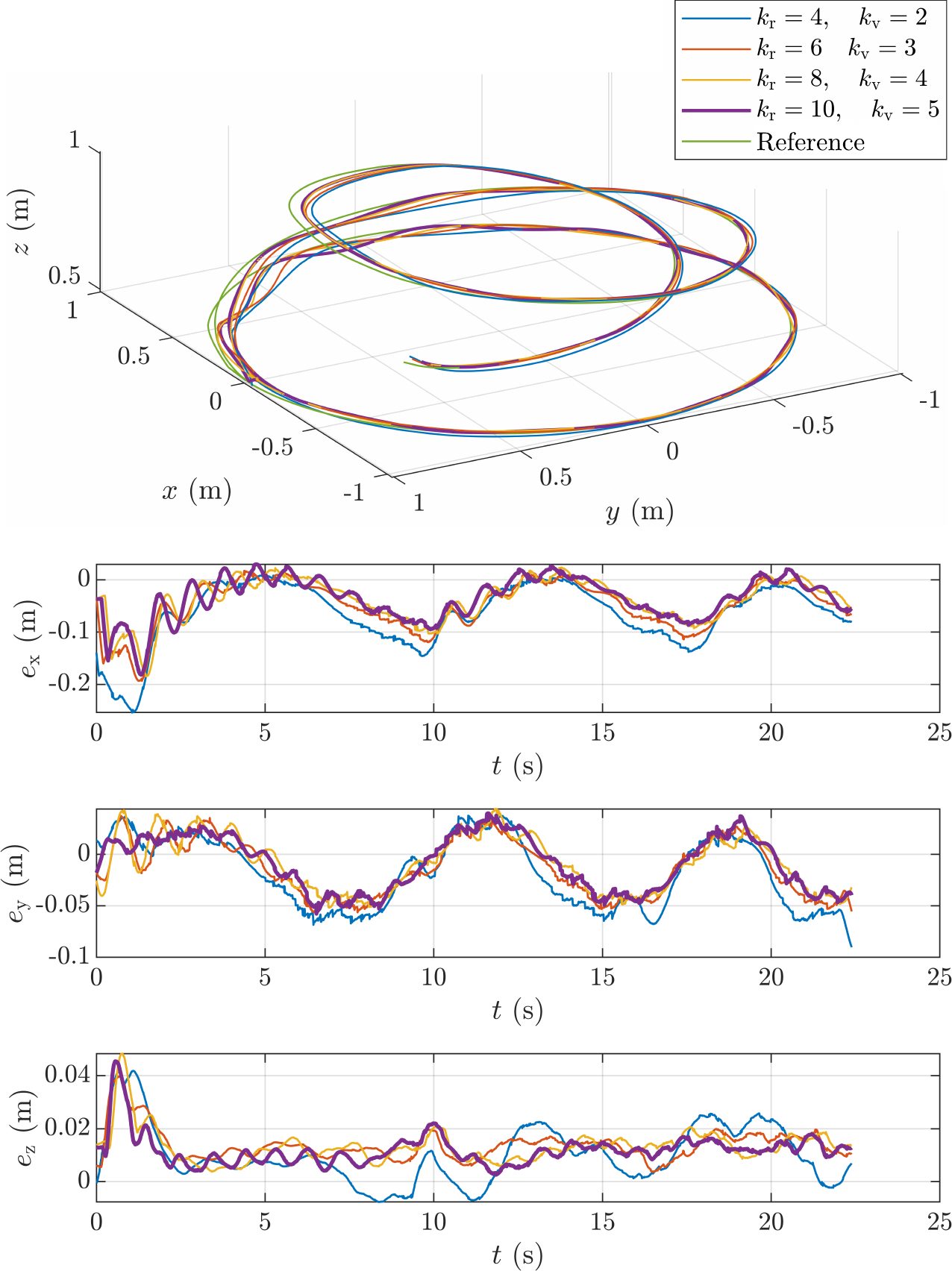}
    \caption{Trajectory tracking during flight experiments with 4 different combinations of control parameters. The upper plot shows the spatial trajectory and the lower plot shows the tracking errors ($e_\mathrm{r} = [e_\mathrm{x}\ e_\mathrm{y}\ e_\mathrm{z}]^\top$).}
    \label{fig:tracking}
\end{figure}

Stable attitude control makes it possible to experiment with position control, for which the control law is given by \eqref{eq:geomforce} with parameters $k_\mathrm{r}, k_\mathrm{v} \in \mathbb{R}$. In this case, the most important property is accurate trajectory tracking, which can be evaluated along a prescribed reference trajectory. To tune the parameters, we have specified a modified helical reference, shown in the upper plot of Figure \ref{fig:tracking} marked in green. Measurements were performed for 4 parameter combinations with the following numerical values:
\begin{equation}
    \begin{bmatrix}
        k_\mathrm{r} \\ k_\mathrm{v} 
    \end{bmatrix} \in \left\{ \begin{bmatrix}
        4 \\ 2
    \end{bmatrix}, \begin{bmatrix}
        6 \\ 3
    \end{bmatrix}, \begin{bmatrix}
        8 \\ 4
    \end{bmatrix}, \begin{bmatrix}
        10 \\ 5
    \end{bmatrix}\right\}.
\end{equation}
From initial tests, we found that using the ratio $k_\mathrm{r}/k_\mathrm{v} = 2$ provides good results, therefore we have fixed it in the 4 cases discussed. The measurement results are shown in Figure \ref{fig:tracking}, where the spatial trajectories and the tracking error terms are displayed. From the figure, it can be seen that the absolute value of the tracking error decreases as the value of the control parameters are increased. This observation was also investigated numerically by evaluating the root mean square of each error term, calculated as follows:
\begin{figure}
    \centering
    \includegraphics[width=8cm]{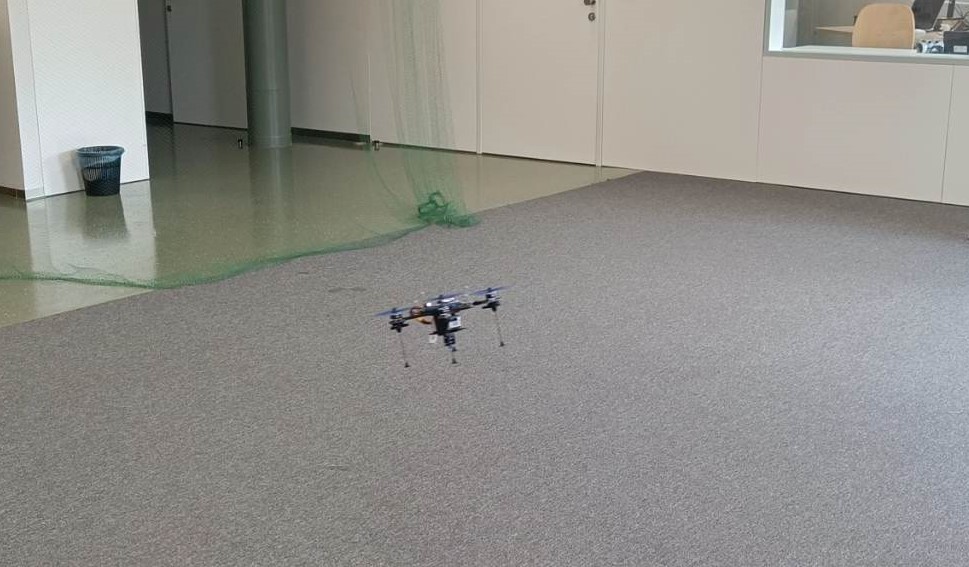}
    \caption{Snapshot during a flight experiment to determine the position control parameters.}
    \label{fig:bb_flying}
\end{figure}
\begin{equation}
    x_\mathrm{RMS} = \sqrt{\frac{1}{N}\sum_{i=1}^N |x_i|^2},
\end{equation}
where $N$ is the dimension and $x_i$ are the elements of the vector $x$. 
As it is shown in Table \ref{tab:rmse}, the minimum root mean square error signals are obtained for parameters $k_\mathrm{r} = 10, k_\mathrm{v} = 5$, which is in line with the observations based on Figure \ref{fig:tracking}. A video recording of the experiment is also available at \url{https://youtu.be/9r_j21rd800}.

\begin{table}[!b]
    \centering \footnotesize
    \caption{The root mean square of the tracking errors, the unit of each value is meter.}
    \begin{tabular}{c|c|c|c|c}
       $k_\mathrm{r}$   &  $k_\mathrm{v}$ & $e_\mathrm{x,RMS}$ & $e_\mathrm{y,RMS}$ & $e_\mathrm{z,RMS}$ \\ \hline \hline
        4 & 2 & 0.0878 & 0.0384 & 0.0147\\
        6 & 3 & 0.0662 & 0.0301 & 0.0151\\
        8 & 4 & 0.0562 & 0.0281 & 0.0147\\
        \textbf{10} & \textbf{5} & \textbf{0.0533} & \textbf{0.0275} & \textbf{0.0133}
    \end{tabular}
    \label{tab:rmse}
\end{table}

\pagebreak
\phantom{a}
\pagebreak
\section{Conclusion}
\vspace{-2mm}
The aim of this work was to develop a procedure to perform agile maneuvers with high tracking performance as quickly as possible on a new quadcopter design with unknown parameters. The advanced motion control algorithms are based on the dynamic model, therefore we first derived the equations of motion of the quadcopter. In order to enable the drone to perform high-speed maneuvers, a nonlinear geometric control algorithm was presented to ensure stability over the full operating domain of the vehicle.

In order to implement the presented algorithms on a real drone, it was necessary to identify the model parameters, which was done partially by CAD modeling and partially through measurements. Finally, flight experiments have been performed to determine the control parameters and it has been shown that the proposed procedure can achieve accurate trajectory tracking with an average tracking error of a few centimeters.

\vspace{-3mm}
\section*{Acknowledgements}
The authors thank Botond Gaál for his help during the flight experiments. This research was supported by the European Union within the framework of the National Laboratory for Autonomous Systems (RRF-
2.3.1-21-2022-00002) and by the Eötvös Loránd Research Network (grant number: SA-77/2021).

\vspace{-3mm}
\bibliographystyle{ieeetr}
{\footnotesize
\bibliography{reference}}

\begin{thebibliography}{10}

\bibitem{argentim2013}
L.~M. Argentim, W.~C. Rezende, P.~E. Santos, and R.~A. Aguiar, ``{PID, LQR and
  LQR-PID} on a quadcopter platform,'' in {\em Proc. of the International
  Conference on Informatics, Electronics and Vision}, pp.~1--6, 2013.

\bibitem{Lee2010}
T.~Lee, M.~Leok, and N.~H. McClamroch, ``Geometric tracking control of a
  quadrotor {UAV} on {SE}(3),'' in {\em Proc. of the 49th IEEE Conference on
  Decision and Control}, pp.~5420--5425, 2010.

\bibitem{Antal2022}
P.~Antal, T.~Péni, and R.~Tóth, ``Nonlinear control method for backflipping
  with miniature quadcopters,'' {\em IFAC-PapersOnLine}, vol.~55, no.~14,
  pp.~133--138, 2022.
\newblock 11th IFAC Symposium on Intelligent Autonomous Vehicles.

\bibitem{Antal2022_2}
P.~Antal, T.~Péni, and R.~Tóth, ``Backflipping with miniature quadcopters by
  {G}aussian {P}rocess based control and planning.'' arXiv: 2209.14652, 2022.

\bibitem{mahony2012}
R.~Mahony, V.~Kumar, and P.~Corke, ``Multirotor {Aerial} {Vehicles}:
  {Modeling}, {Estimation}, and {Control} of {Quadrotor},'' {\em IEEE Robotics
  \& Automation Magazine}, vol.~19, no.~3, pp.~20--32, 2012.

\bibitem{Bisheban2019}
M.~Bisheban, {\em Geometric Control of a Quadrotor Unmanned Aerial Vehicle in
  Wind Fields}.
\newblock PhD thesis, George Washington University, 2019.

\bibitem{lee2009}
D.~Lee and S.~Sastry, ``Feedback linearization vs. adaptive sliding mode
  control for a quadrotor helicopter,'' {\em International Journal of Control,
  Automation and Systems}, vol.~7, pp.~419--428, 06 2009.

\bibitem{nieuwstadt1996}
M.~J. Van~Nieuwstadt and R.~M. Murray, ``Real {Time} {Trajectory} {Generation}
  for {Differentially} {Flat} {Systems},'' {\em International Journal of Robust
  and Nonlinear Control}, vol.~8, no.~11, pp.~995--1020, 1998.

\bibitem{mellinger2011}
D.~Mellinger and V.~Kumar, ``Minimum snap trajectory generation and control for
  quadrotors,'' in {\em Proc. of the IEEE International Conference on Robotics
  and Automation}, pp.~2520--2525, 2011.

\bibitem{Forster}
J.~Förster, ``System identification of the {C}razyflie 2.0 nano
  quadrocopter,'' {Bachelor's Thesis}, ETH Zurich, Zurich, 2015.

\bibitem{Mustapa2014}
Z.~Mustapa, S.~Saat, S.~H. Husin, and T.~Zaid, ``Quadcopter physical parameter
  identification and altitude system analysis,'' in {\em 2014 IEEE Symposium on
  Industrial Electronics and Applications (ISIEA)}, pp.~130--135, 2014.

\end{thebibliography}

\end{document}